\pdfoutput=1
\documentclass[11pt]{article}

\usepackage{EMNLP2022}

\usepackage{times}
\usepackage{latexsym}
\usepackage{CJKutf8}

\usepackage[T1]{fontenc}
\usepackage[utf8]{inputenc}

\usepackage{microtype}

\usepackage{inconsolata}

\usepackage{graphicx}

\usepackage{xurl}

\title{Unsupervised Tokenization Learning}

\author{Anton Kolonin \\ Aigents \\ SingularityNET Foundation \\ Novosibirsk State University \\ \texttt{akolonin@gmail.com}
         \And
        Vignav Ramesh \\ Harvard University \\ SingularityNET Foundation \\ \texttt{vignavramesh@college.harvard.edu}
}

\begin{document}
\maketitle
\begin{abstract}
In the presented study, we discover that the so-called “transition freedom” metric appears superior for unsupervised tokenization purposes in comparison to statistical metrics such as mutual information and conditional probability, providing F-measure scores in range from $0.71$ to $1.0$ across explored multilingual corpora. We find that different languages require different offshoots of that metric (such as derivative, variance, and “peak values”) for successful tokenization. Larger training corpora do not necessarily result in better tokenization quality, while compressing the models by eliminating statistically weak evidence tends to improve performance. The proposed unsupervised tokenization technique provides quality better than or comparable to lexicon-based ones, depending on the language.
\end{abstract}

\section{Introduction}

Unsupervised language learning, framed as a problem of language modeling based on unannotated corpora, has attracted great attention in recent years, having achieved significant results with transformer-based models such as BERT and GPT that rely on deep neural networks (DNNs) \citep{1,2}. At the same time, the idea of unsupervisedly learning a language grammar represented “interpretably” via a formal grammar such as Link Grammar has been suggested by \citet{3}. \citet{4} proposed yet another approach for the problem: using so-called “deep patterns” with hierarchical “symbolic” grammatical pattern structures learned from texts as a way to model grammars and ontologies for natural languages suiting a wide range of practical applications. Further studies on this path performed by \citet{5,6} have indicated the possibility of learning grammars as well as domain ontologies given high-quality parse trees of texts obtained from unannotated training corpora. Unfortunately, the critical part of the pipelines described in the aforementioned studies was the unsupervised generation of the parses, which turned out to be low-quality by virtue of being based on simple “minimum spanning trees” either based on mutual information (MI) \citep[see][]{7} or “contextual information” \citep[see][]{6} extracted from a BERT-based deep learning model \citep{1}. Still further studies by Ramesh and Kolonin have demonstrated the possibility of building different natural language processing (NLP) applications based on a language model represented by a formal grammar (Link Grammar in the explored cases) \citep{8,9,10}. Throughout the course of these studies, the concept of “interpretable natural language processing” (INLP) has been introduced to indicate the domain of NLP explorations involving both the learning of language models represented in an interpretable form and the application of these models to different tasks such as text segmentation, language generation, and question answering. 

Yet another problem which has its place in the case of conventional language model learning based on DNNs \citep[see][]{1,2} as well as in relation to interpretable unsupervised language learning \citep[see][]{4,5,6} is tokenization. In most cases, tokenization is based on predetermined rules and dictionaries, which does not quite fit the “grand plan” of completely unsupervised language learning from scratch with no prior knowledge of the language, including its lexicon and punctuation \citep{3}. Thus, the objective of our proposed study is to evaluate the possibility of learning the sets of tokens representing both punctuation and lexicons without any prior knowledge regarding the language, so that the set of valid combinations of letters or characters specific to punctuation marks or valid lexical entities such as words for a language is learned along with the tokenization process.

Starting points were found in works of \citet{11} and \citet{12}, who explored the possibility of unsupervised segmentation applied to different languages and domain-specific literature.

The former work \citep[see][]{11} provides an exhaustive overview of different tokenization techniques applied to different languages, exploring different methods and metrics. Unfortunately, the $F_1$ scores reported in this work for completely unsupervised tokenization based on statistical measures appear not high enough, so we further follow this approach in order to outperform these scores on the set of languages relevant and available to us—English, Russian, and Simplified (Mainland) Chinese—focusing on unsupervised tokenization only.

The latter work \citep[see][]{12} focuses on unsupervised tokenization based on statistical measures such as conditional probability (CP) as well as introduces the so-called “freedom of transition” (we henceforth call it “transition freedom” or TF) metric, which appears fundamentally consistent with the notion of “free energy” suggested by \citet{13} as a key for an artificial intelligence concept. TF in the context of \citeposs{12} work corresponds to the number of symbolic states (characters, letters, or N-grams) that follow or precede the current state. Then, a sharp increase of the TF level along the temporal sequence of states might correspond to a loss of \citeposs{13} “equilibrium,” and so “tokens” might be considered as chains of states resting in conditions of mutual equilibrium framed with transitions, with loss of this equilibrium marked by the TF level bursts. Furthermore, we explore both statistical measures and TF metrics, finding the latter substantially more practical. In particular, we explore different metrics based on CP and TF such as derivative, variance, and “peak values” as introduced in \citeposs{12} work (which indicate expressed local maximums on the derivative curve along the text being tokenized).  

Interestingly, \citet{11} writes that, “Given that the human ability to successfully read any natural language provides an existence proof that a generalized segmentation system (as implemented in the human mind) is possible, it is reasonable to investigate the feasibility of a language-agnostic segmentation system that could be easily integrated into larger natural language processing systems.” Extending this statement, we anticipate that advances in this area could be also beneficial to deal with any sequential data such as flows of events and states in experiential or reinforcement learning. In particular, the “global feedback” concept suggested in \citeposs{14} work demonstrates good learning rates in cases when the cognitive schema leading to the feedback or reward can be reliably associated with entire sequences of preceding actions, which is difficult to deal with in existing reinforcement learning frameworks. The ability to segment sequences of cognitive experiences unsupervisedly might potentially advance research on experiential and reinforcement learning with delayed reward or with no explicit feedback in any subject domain beyond NLP. The importance of the latter goal is also outlined in \citeposs{15} work, where it was stated that the “discovery of reusable sub-routines simplifies decision-making and planning in complex reinforcement learning problems.”

As will be presented further, we find the TF to be superior over MI \citep[see][]{5, 6, 7, 11} and CP \citep[see][]{12} for the task of unsupervised text segmentation (tokenization). We find that the English and Russian languages require one specific way of handling the TF (variance) while Chinese requires a slightly different way (derivative-based “peak values”) for the same purpose. Tokenization quality for English and Russian may have $F_1$ scores as high as $0.96$-$1.0$ depending on training and testing corpora, while for Chinese the best score is $F_1=0.71$ with precision of lexical word discovery reaching $0.92$. Larger training corpora do not necessarily result in better tokenization quality, while compressing the models by eliminating statistically weak evidence typically improves the quality. Unsupervised TF-based tokenization provides quality that is the same as or better than lexicon-based tokenization for English and Russian, while for Chinese it appears to be the opposite (as could be anticipated); however, the precision of lexicon discovery for Chinese using TF-based tokenization appears close to reference tokenization.

\section{\label{data}Data Sets}

We have used different training data sets for three different languages (English, Russian, and Chinese), while the same parallel corpus has been used for testing.

For English training corpora we have used the Brown (\url{http://www.sls.hawaii.edu/bley-vroman/brown_nolines.txt}), Gutenberg (\url{https://www.gutenberg.org}) Children, and Gutenberg Adult collections, as well as mixed collections such as Gutenberg Children and Adult blended together and all three corpora blended together. The sizes of the above corpora are $6$M, $29$M, and $140$M, respectively. 

For Russian training corpora we have used the RusAge collection (\url{https://www.kaggle.com/datasets/oldaandozerskaya/fiction-corpus-for-agebased-text-classification}) as two separate pieces: Test ($141$M size) and Previews ($825$M size), each used as an independent training corpus.

For Chinese training corpora we have used the CLUE Benchmark News 2016 dataset (\url{https://github.com/brightmart/nlp_chinese_corpus}), which contains two pieces: Train and Validation. Each piece was used as an individual training dataset. The raw data encoded in JSON format have been processed so that \texttt{title}, \texttt{desc}, and \texttt{content} fields were extracted individually and each of the three fields was saved on a separate line in the text file used as input for further processing. After such preprocessing, we obtained an $8,500$M-size training dataset and a $270$M-size Validation dataset. 

For the test corpus across all three languages above, we have used a parallel Chinese/English/Russian corpus of $100$ multi-sentence statements within the financial domain, as derived from the dataset released by Magic Data (\url{https://magichub.com/datasets/chinese-english-parallel-corpus-finance}). The original corpus is parallel Chinese/English, but the Russian version of all $100$ statements have been added with the help of Google Translate, with Chinese proper names manually replaced with Russian or English proper names used in the appropriate subject domain context.

English and Russian reference lexicons have been obtained from Aigents/Pygents open source project on: English obtained from (\url{https://raw.githubusercontent.com/aigents/aigents-java/master/lexicon_english.txt}), Russian (\url{https://raw.githubusercontent.com/aigents/aigents-java/master/lexicon_russian.txt}).

A few different Chinese lexicons were obtained for reference: Chinese Lexical Database, or CLD (\url{http://www.chineselexicaldatabase.com/download.php}) \citep[see][]{16};  BLCU Chinese Corpus, or BLC (\url{https://www.plecoforums.com/threads/word-frequency-list-based-on-a-15-billion-character-corpus-bcc-blcu-chinese-corpus.5859}); and SUBTLEX-CH (\url{http://crr.ugent.be/programs-data/subtitle-frequencies/subtlex-ch}).

\section{Exploration Methodology}

\subsection{Overview}

Our study involved the following phases:

\begin{itemize}

\item Models were trained on each training corpora across all three languages.

\item Tokenization was performed for each of the languages with the models created in the previous phase, using different training corpora with different metrics and hyperparameters as will be discussed further. The same parallel test corpus was used for each language. While performing tokenization, $F_1$ scores were evaluated for every set of hyperparameters and selected metrics, comparing the tokenization outputs with the outputs of a “standard” lexicon-based reference tokenizer. At this point, the corpora and sets of hyperparameters leading to the best $F_1$ scores per language have been identified.

\item The tokenization configurations corresponding to “winning” (superior) $F_1$ scores were evaluated in comparison to the reference lexicon-based tokenizer specific to each of the three languages.

\item The winning configurations were evaluated based on precision of lexicon discovery. This process included determining the fraction of tokens identified by the best unsupervised tokenizer setup that actually correspond to entries in reference lexicon dictionaries for each language.

\end{itemize}

\subsection{\label{sec:s1}Model Structure and Building}

Each of the models created for a corpus was represented by three pieces, based on N-grams with $N$ in range from $1$ (unigrams or individual characters/letters) to $N_{max}$ (up to $7$, according to discussion in \citeposs{12} work), with the latter being one of the hyperparameters discussed further. These pieces are described below.
\begin{itemize}
\item N-gram frequencies or counts of N-grams experienced through the corpus.
\item Counts of all N-grams appearing after every specific N-gram (we call them “forward transitions”).   
\item Counts of all N-grams appearing before every specific N-gram (we call them “backward transitions”).
\end{itemize}
The model building process has been applied to corpus data on a line-by-line basis according to the original text layout of the corpora, without any other preprocessing.

For transition counts, two different models were built for every language corpus. First, there was \texttt{N-gram-to-symbol} counts, where the number of single symbols (unigrams) following or preceding every possible N-gram were counted. Second, there was \texttt{N-gram-to-N-gram} transitions, where the number of N-grams following or proceeding a given N-gram were counted ($N$ being the same). Preliminary studies on English corpora run at the beginning of our exploration have shown inferior performance of the latter kind of models, so further studies involved the \texttt{N-gram-to-symbol} models only.

The value of $N$ varied from $1$ to $7$ for each language except Chinese, where $N_{max}=3$ for the smaller Validation dataset and $N_{max}=2$ for the larger Training dataset due to memory restrictions of 32G RAM which made it impossible to process larger models for Chinese corpora. 

The described model of a language based on given corpora can be represented as a bidirected graph, with transitions on graph edges pointing both forward and backward independently. Every symbolic unit was involved in multiple overlaid subgraphs due to multiple contexts represented by embedding the same N-gram in multiple transitions on the graph as well as by embedding N-grams of lower rank into multiple N-grams of higher rank. The bidirected graph was weighted by frequency counts associated with vertices corresponding to N-grams as well as with edges corresponding to transitions. The same graph might be viewed in three ways: as an excessive container including a graph-based grammatical model expressed in a formal grammar such as Link Grammar \citep{3,5,6}; as a bottom layer of the heterarchical system of “deep patterns” which can be used to infer higher-level abstractions \citep{4}; or as a set of interconnected symbolic “instances” underlying an abstract higher-level language model consisting of interconnected symbolic “invariants” corresponding to parts of speech \citep{17}.

\subsection{\label{s2}Tokenization Methods and Metrics}
The following tokenization methods and respective metrics were used:
\begin{itemize}
\item Greedy aggregation of symbols into tokens according to the mutual information computed for pairwise symbol associations, as described by \citet{7} and \citet{11}. This did not work well in the initial cursory study on English corpora (proper English words were systematically broken into pieces), so it was not further considered as a tokenization approach.  
\item Probability (P) of an N-gram. N-grams with lexicon-wise probabilities above certain thresholds serve as delimiters breaking the stream of symbols into tokens.
\item Conditional Probability (CP) computed on derivatives of \texttt{N-gram-to-N-gram} transitions in both forward and backward directions as described by \citet{11} and \citet{12}, with local maximums on \texttt{N-gram-to-N-gram} transitions corresponding to token breaking points. 
\item CP variance, the difference between the CP and its mean value for a given input sequence.
\item Transition Freedom (TF), the number of possible transitions on forward or backward model graph traversal at a specific N-gram according to methodology described by \citet{12}, with values exceeding the threshold breaking the stream of symbols into tokens. 
\item TF variance, defined as the difference between the TF and its mean value for a given input sequence.
\item TF derivative in both directions, with local maximums on \texttt{N-gram-to-N-gram} transitions corresponding to token breaking points.
\item TF “peak values” defined in \citeposs{12} work as a value of TF derivative on the previous transition minus the value of TF derivative on the following transition. Such peak values outline sharp positive extremums of the TF curve along the processed sequence of N-grams, indicating the token boundaries. They can be interpreted as negative second derivatives shifted one point back.
\item Lexicon-based tokenization in “greedy” mode, so that either the longest or most frequent token entry present in the language-specific lexicon dictionary is identified as a next token when traversing the input text forward from left to right (blending the two criteria of length and the logarithm of frequency has also been explored). As this is not an unsupervised approach, being based on a pre-created lexicon, this tokenization was used only for reference.
\item Reference “hardcoded tokenizer” used to assess the $F_1$ scores of the unsupervised tokenizer. In the cases of English and Russian, it was a simple text splitter based on white spaces with quotes, brackets, periods, commas, semicolons, and other punctuation symbols detached from the split token sequence. For Chinese, it was the Jieba Tokenizer, which uses a combination of hardcoded rules, built-in dictionaries, and probabilistic measures \citep{18}. 
\end{itemize}

\begin{figure}
  \includegraphics[width=0.49\textwidth]{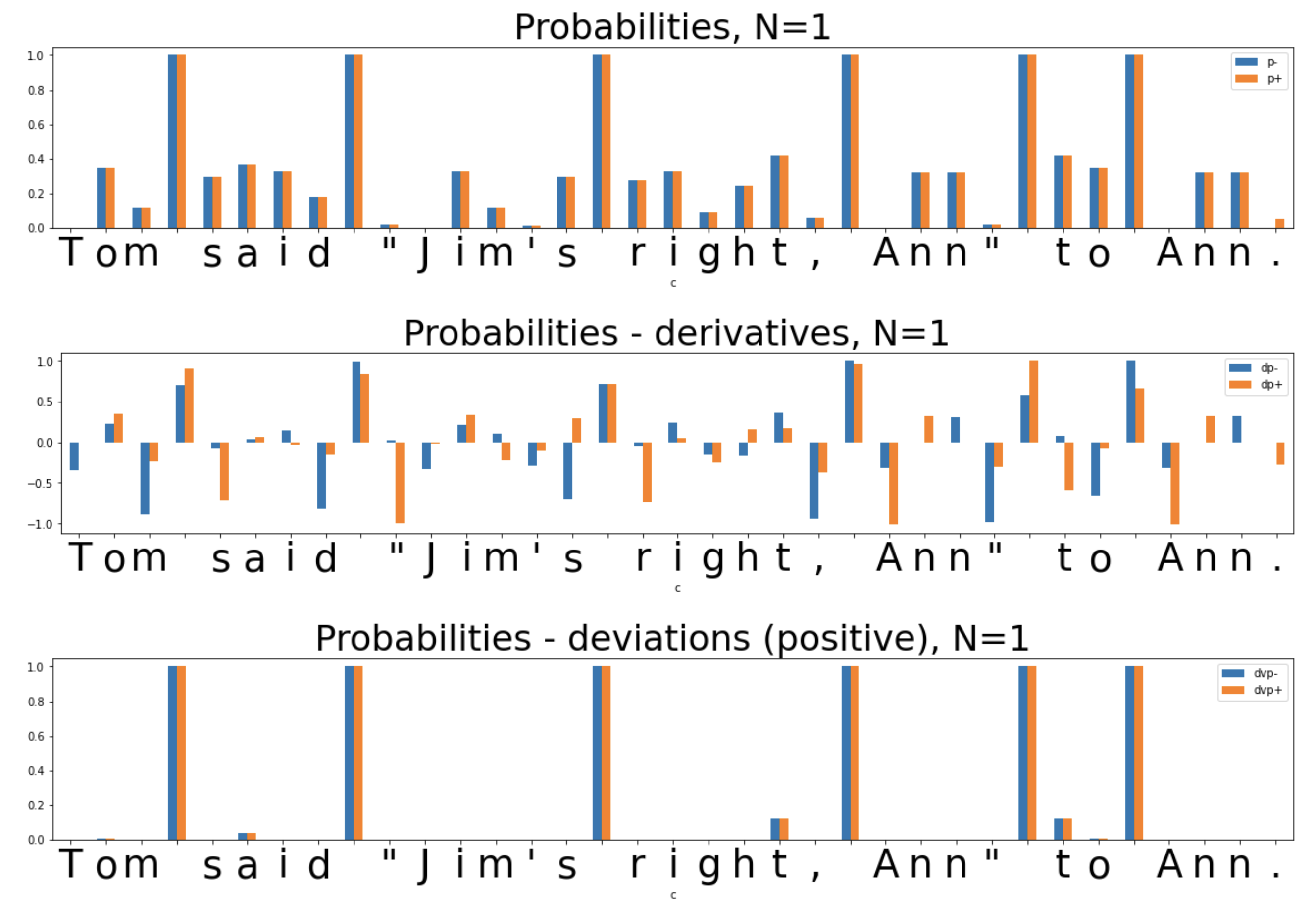}
  \caption{Using probabilities (p) and derived metrics such as variance (dvp) and derivatives in forward (dp+) and backward (dp-) traversals. It is clearly seen that punctuation marks cannot be isolated from words.}
\end{figure}

\begin{figure}[!ht]
  \includegraphics[width=0.49\textwidth]{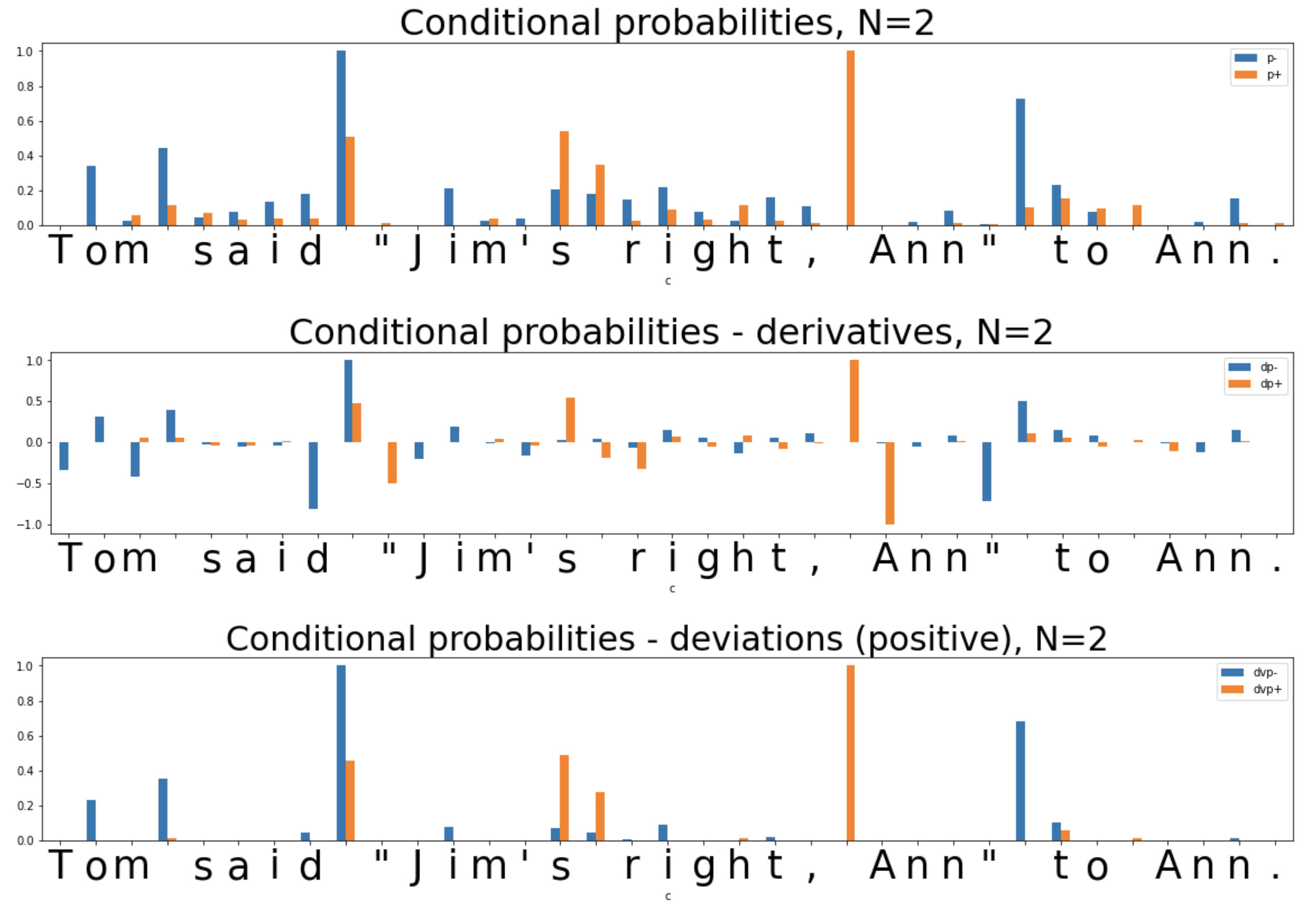}
  \caption{Using conditional probabilities (p) and derived metrics such as derivative in forward (dp+) and backward (dp-) transitions and variance (dvp+ and dvp-, respectively) computed on bigrams. It is clearly seen that punctuation marks cannot be isolated from words, and some of the words are disassembled into pieces.}
\end{figure}

\begin{figure}[!ht]
  \includegraphics[width=0.49\textwidth]{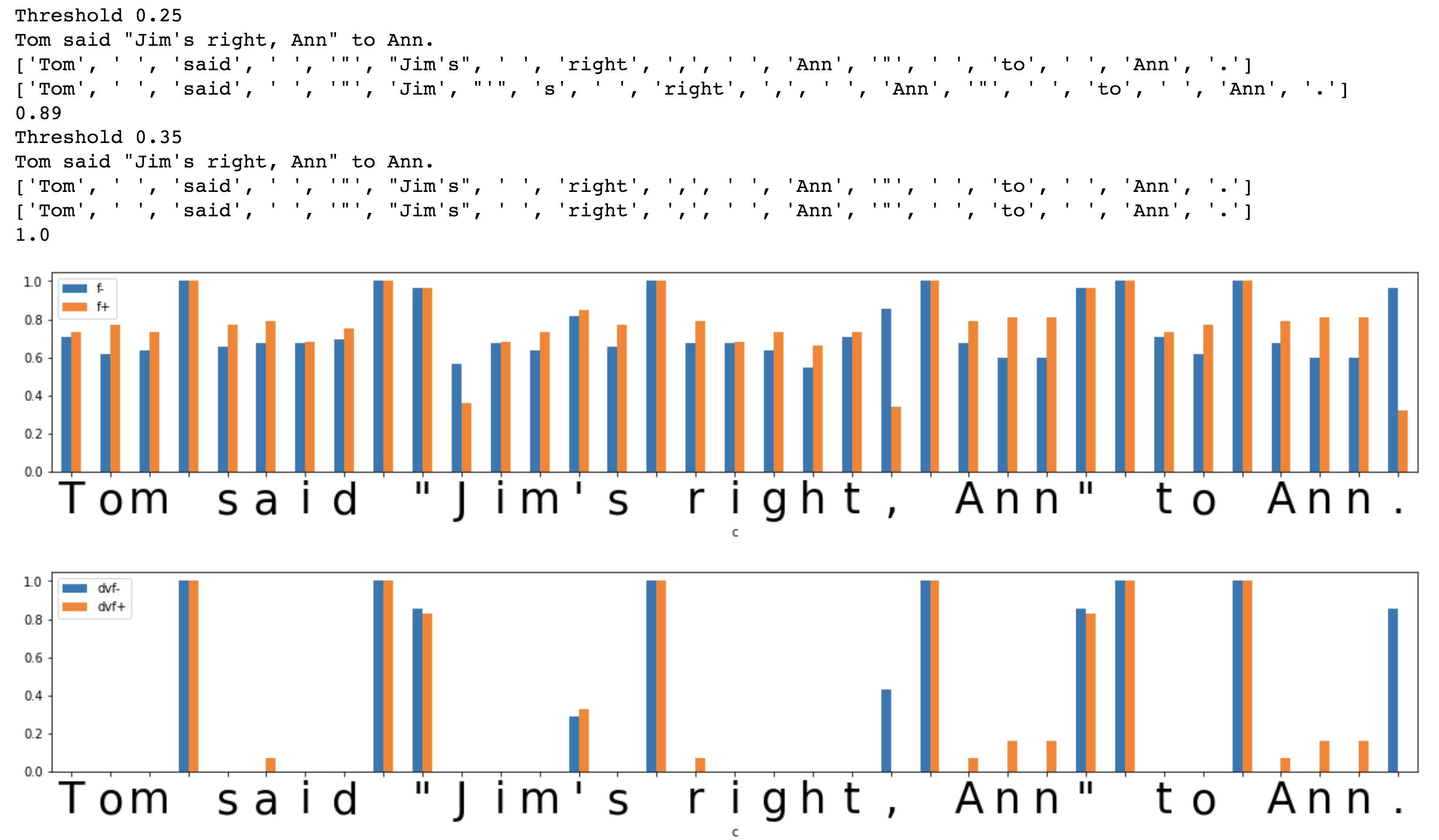}
  \caption{Using transition freedoms in forward (f+) and backward (f-) directions and their variances (dvf+ and dvf-) computed on unigrams. All words as well as punctuation marks are identified clearly, with threshold values of $0.25$ and $0.35$.}
\end{figure}

For all methods relying on CP and TF metrics above, two alternative ways of identifying token boundaries are possible. First, as suggested by \citet{12}, the “mean” metric is computed on forward and backward traversals over the sequence of N-grams referring to corresponding subgraphs in the model. Second, the token break is identified as a metric derived from either P, CP, or TF exceeding the threshold on either forward or backward transitions along the text (i.e.,  the “max” was used instead of the “mean”). Cursory checks across corpora have shown that the “mean” method is not quite reliable compared to the “max” alternative, so the latter method was used in the studies presented below.

\begin{figure}[!ht]
  \includegraphics[width=0.49\textwidth]{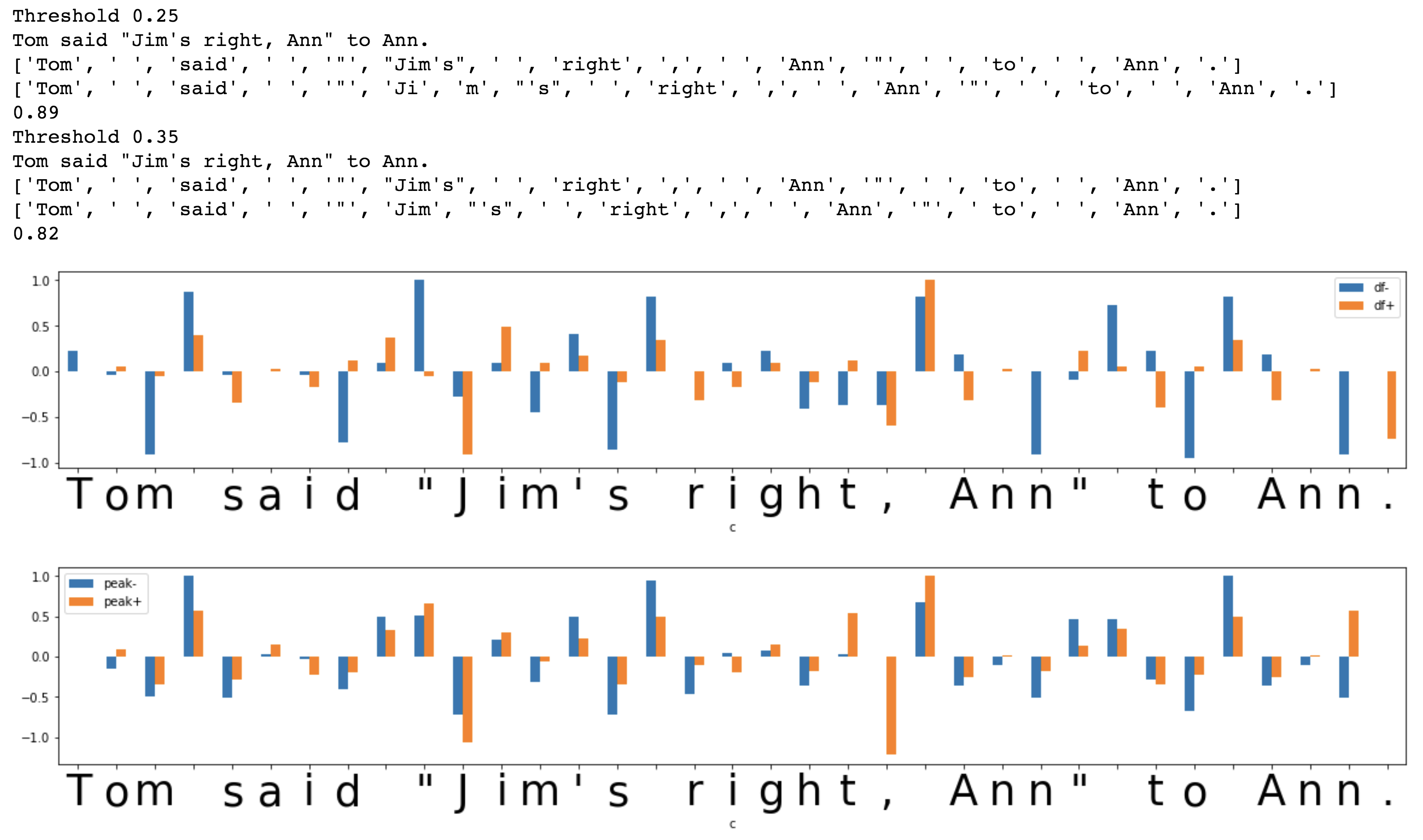}
  \caption{Using TF derivatives in forward (df+) and backward (df-) directions and their “peak values” (peak+ and peak-) computed on unigrams. Some words are not identified clearly and punctuation marks are not separated. Threshold values are $0.25$ and $0.35$.}
\end{figure}

\begin{figure}[!ht]
  \includegraphics[width=0.49\textwidth]{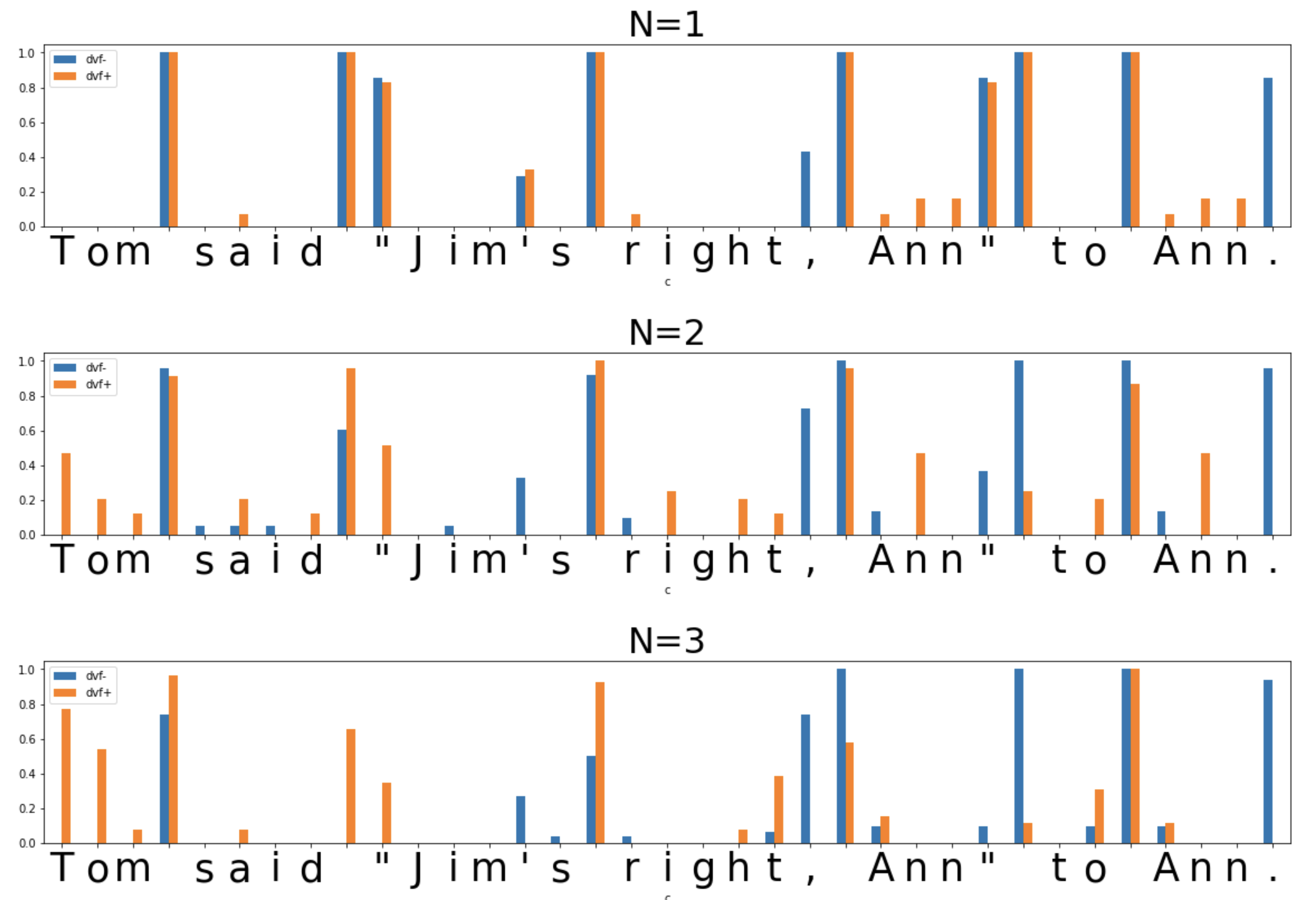}
  \caption{Variances of transition freedoms in forward (dvf+) and backward (dvf-) directions with N-grams of varying $N$-values. It is clearly apparent that $N=1$ allows for the most accurate identification of whitespaces as well as punctuation marks.}
\end{figure}

In trying to reach superior $F_1$ scores during the studies, we also explored if it would help to “compress” the model by eliminating edges on the graph with weights below a certain threshold measured relative to the maximum N-gram or transition frequency in the local subgraph segment—that is, if a model derived from the raw model with removal of all low-frequency N-grams and low-frequency transitions for any given N-gram can increase $F_1$.

\subsection{Tokenization $F_1$ Score and Precision of Lexicon Discovery}

$F_1$ scores were calculated to evaluate our unsupervised tokenizer by comparing its performance to that of the reference tokenizer based on “hardcoded” logic. The scores were computed based on a non-unique set of tokens with counted occurrences (e.g., each repetition of the determiner “the” in a tokenized text is considered separately).

The other evaluative metric we used was the unsupervised tokenizer’s capacity to discover lexical entities for an unknown language, called precision of lexicon discovery. This metric was evaluated as the ratio of all tokens found in an input text present in a reference lexicon dictionary to the total number of tokenized entries.

\subsection{\label{35}Tokenization Hyperparameters}

As mentioned above, there were a few hyper-parameters explored in further experiments on unsupervised tokenization in a unified manner across all three languages studied:
\begin{itemize}
\item Tokenization metric: use either P/CP or TF as a base metric and then use either the base metric value itself or an offshoot of it such as variance, derivative, or “peak value.”
\item The combination of $N$ ranks used to perform model graph traversal and the “mean” metric computation based on a specified subset of N-grams. We have explored every possible individual value of $N$ as well as arbitrary combinations of $N$-values.     
\item Model compression threshold used to remove low-frequency N-grams (corresponding to vertices and transitions between them on the model graph). We have used the following values: $0.0$ (corresponding to no compression at all), $0.0001$, $0.001$, $0.01$, and $0.1$.
\item Tokenization metric threshold: the value of a metric exceeding this level would correspond to a token boundary. We have used the following values: $0.0001$, $0.0005$, $0.001$, $0.005$, $0.01$, $0.02$, $0.05$, $0.1$, $0.2$, $0.3$, $0.4$, $0.5$, $0.6$, $0.7$, $0.8$, $0.9$.
\end{itemize}

 “Grid search” was employed to find the best configuration of these four hyperparameters—that is, the setup providing superior $F_1$ scores.

The “winning” configuration of the hyperparameters obtained for the full test set of $100$ sentences per language was validated as follows: independent splits of the test set into two sets of $50$ sentences obtained nearly the same results for the same combinations of hyperparameters without changing the configuration.

\section{Experimental Results}

\subsection{\label{41}English}

We obtained a maximum tokenization $F_1$ score of $0.99$ using the TF variance metric after training on the smallest Brown corpus; $N=1$ (unigrams); model compression thresholds of $0.0001$ and $0.001$; and tokenization thresholds of $0.4$ and $0.5$.  Using larger or blended corpora allowed for $F_1$ scores above $0.93$ but below $0.99$ with similar hyperparameter configurations.

\begin{figure}
  \includegraphics[width=0.49\textwidth]{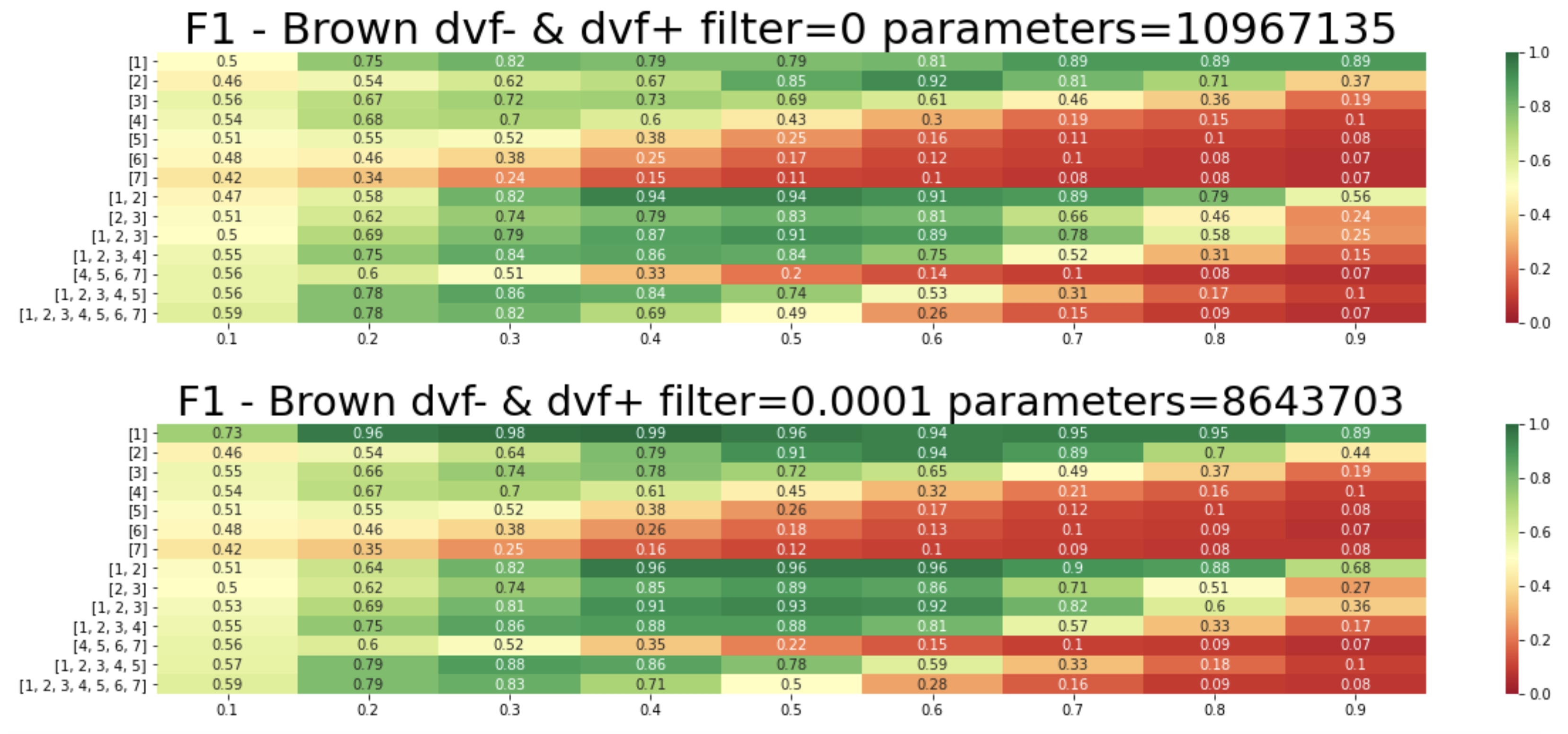}
  \caption{Heat-maps rendering $F_1$ scores obtained for unsupervised tokenization after training on the Brown corpus with no model compression (top) and model compression with a threshold of $0.0001$ (bottom) with different combinations of $N$ (vertical axes) and different tokenization thresholds (horizontal axes). It is seen that the highest $F_1$ scores above $0.96$ correspond to models compressed with threshold $0.0001$, $N=1$ (unigrams), and tokenization thresholds from $0.3$ to $0.4$. Model parameters are indicated in the plot titles, where each parameter corresponds to the weight or frequency count for either N-grams or transitions between N-grams.}
\end{figure}

Lexicon-based tokenization in “greedy” mode, driven by token lengths, provided the same level of performance with $F_1=0.99$, after having delimiting symbols added to the reference lexicon dictionary.  

Precision of word discovery with unsupervised tokenization turned out to be $0.99$ (after correction for proper English words missed in the reference lexicon dictionary)—a result comparable to reference delimiter-based tokenization ($1.0$). The $0.01$ error was caused primarily by the unsupervised tokenizer’s inability to recognize question marks attached to the ends of words as separate tokens. This expectedly might be solved with larger corpora involving a greater variety of question marks included in different contexts, because all other punctuation marks have been identified correctly as separate tokens.   

\subsection{\label{42}Russian}

We obtained a maximum tokenization $F_1$ score of $1.0$ using the TF variance metric after training on any corpora; $N=1$ (unigrams); a model compression threshold of $0.0001$ for all training corpora (and even no compression at all for smaller corpora); and a tokenization threshold of $0.7$.

Lexicon-based tokenization in “greedy” mode provided a lower level of performance ($F_1=0.94$) due to the words missed in the lexicon, after having delimiting symbols added to the reference lexicon dictionary.  

Precision of word discovery with unsupervised tokenization turned out to be $1.0$ (after correction for proper Russian words missed in the reference lexicon dictionary), equal to that of reference delimiter-based tokenization.

\subsection{\label{43}Chinese}

We obtained a maximum tokenization $F_1$ score of $0.71$ using the TF “peak” metric; $N=2$ (bigrams); model compression thresholds of $0.001$ on larger training corpora; and any tokenization threshold between $0.0$ and $0.05$, inclusive. Unfortunately, we were not able to explore N-grams with $N > 3$ for smaller lexicons and $N>2$ for larger lexicons due to the $32$G memory limit on our model, which was implemented in Python using plain dictionaries for graph model storage. 

Lexicon-based tokenization in “greedy” mode provided a higher level of performance ($F_1=0.83$).

\begin{figure}
  \includegraphics[width=0.49\textwidth]{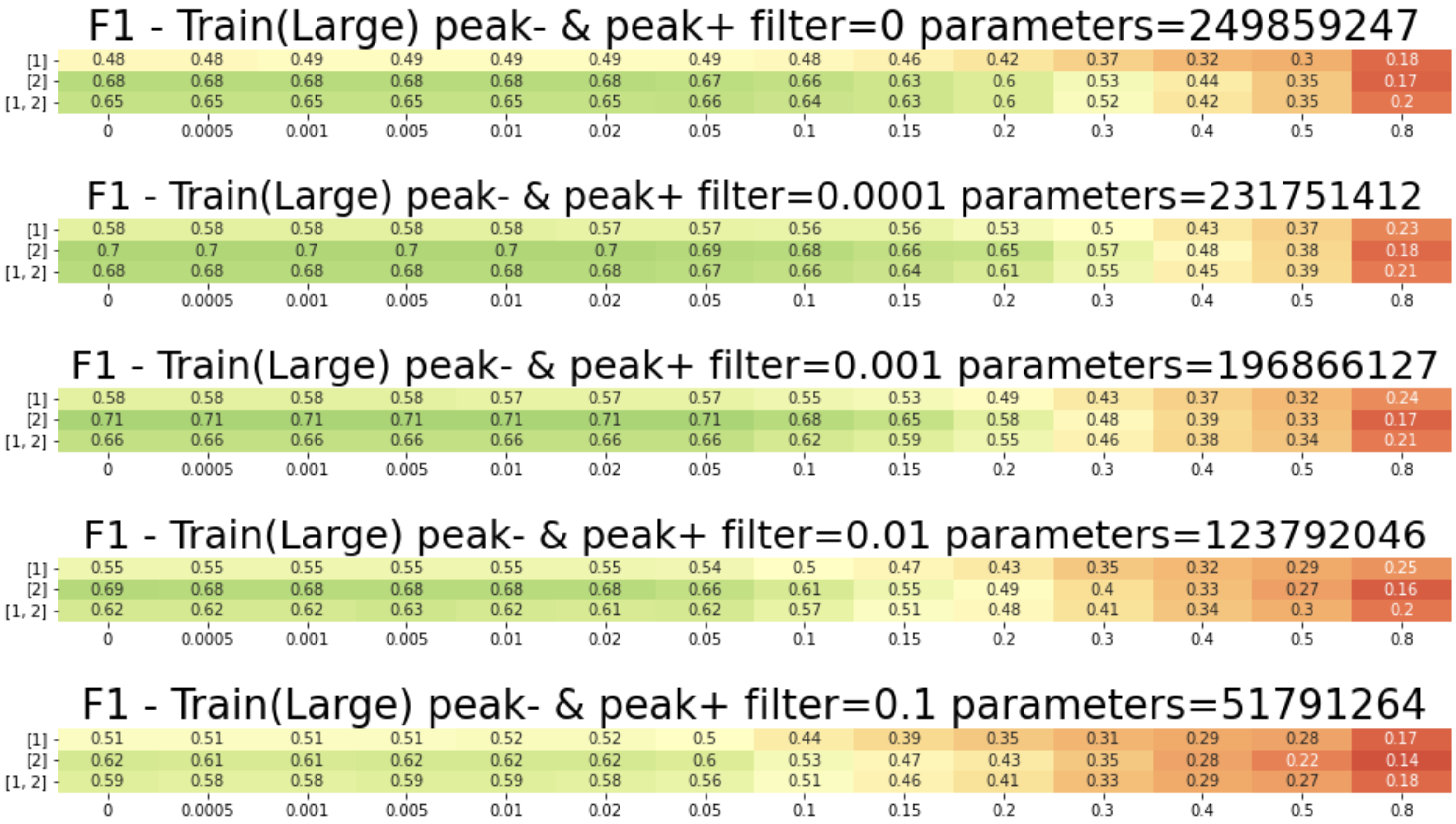}
  \caption{Heat-maps rendering $F_1$ scores obtained for unsupervised tokenization after training on the Chinese CLUE Benchmark News 2016 corpus with no model compression (top) and model compression thresholds from $0.0001$ to $0.1$ (top down) with different combinations of $N$ (vertical axes) and different tokenization thresholds (horizontal axes).}
\end{figure}

\begin{figure}[!ht]
  \includegraphics[width=0.49\textwidth]{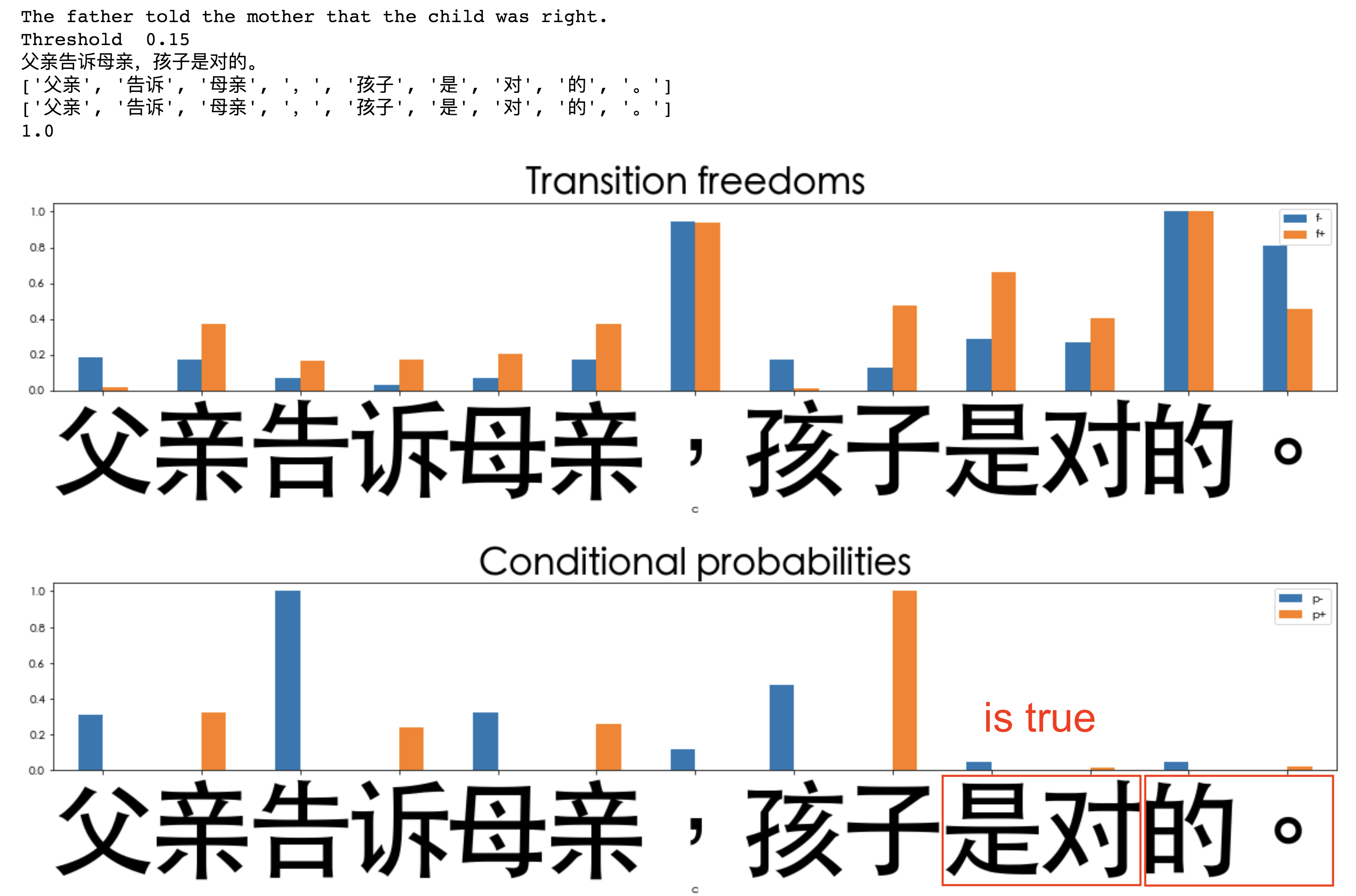}
  \caption{Using transition freedoms in forward (f+) and backward (f-) directions computed on bigrams for Chinese. All proper words and punctuation marks are identified clearly with a threshold value of $0.15$ (top chart). When probabilities (p+ and p- on the bottom chart) computed on bigrams are used, the Chinese period “\begin{CJK*}{UTF8}{gbsn}。\end{CJK*}” is not correctly identified as a separate token (rightmost bounding box at the bottom); however, the “token” containing these two symbols still appears semantically valid in the broader context of the sentence (“is true” makes sense as used above). As with question marks in English, this problem might be solved using richer training corpora with a greater diversity of contexts in which the given punctuation is used.}
\end{figure}

Regardless, it seems that the mistakes made by the tokenizer for Chinese tests did not significantly impact the meaning of the tokenized output, assuming that translations of alternative combinations of symbols looked up in Google Translate were accurate (the authors have minimal firsthand knowledge of the Chinese language).

\begin{table*}
\centering
\begin{tabular}{llcc}
\hline
\textbf{Language}&\textbf{Tokenizer} &\textbf{Tokenization $F_1$} &\textbf{Lexicon Discovery Precision}\\
\hline
English & Freedom-based & \textbf{$\mathbf{0.99}$} & \textbf{$\mathbf{0.99}$} (vs. $1.0$)\\
English  & Lexicon-based & $0.99$ & -\\
% English no spaces & Freedom-based & 0.42 & -\\
% English no spaces & Lexicon-based & 0.79 & -\\
Russian  & Freedom-based & \textbf{$\mathbf{1.0}$} & \textbf{$\mathbf{1.0}$} (vs. $1.0$)\\
Russian  & Lexicon-based & $0.94$ & -\\
% Russian no spaces & Freedom-based  & 0.26 & -\\
% Russian no spaces  & Lexicon-based & 0.72 & -\\
Chinese  & Freedom-based & \textbf{$\mathbf{0.71}$} & \textbf{$\mathbf{0.92}$} (vs. $0.94$)\\
Chinese  & Lexicon-based & $0.83$ & -\\
\hline
\end{tabular}
\caption{\label{table}Summary of the presented research on tokenizers relying on Transition Freedom (“Freedom-based”) or on loaded lexicons (“Lexicon-based”). The last column provides reference numbers for rule-based tokenizers (models based on hardcoded rules)/hybrid tokenizers (models combining hardcoded rules, lexicons, and/or statistical measures such as mutual/conditional probabilities) in parentheses. \textbf{English}: Both tokenization and lexicon discovery are solved with freedom-based tokenizers no worse than with lexicon-based ones ($F_1$ and Precision $=0.99$). \textbf{Russian}: Both tokenization and lexicon discovery tasks are solved better ($F_1$ and Precision $=1.0$) with freedom-based tokenizers than with lexicon-based ones ($F_1=0.94$). \textbf{Chinese}: Tokenization is solved less accurately by freedom-based tokenizers than by lexicon-based ones ($0.71$ vs. $0.83$). However, freedom-based tokenizers perform lexicon discovery relatively well compared to rule-based/hybrid tokenizers ($0.92$ vs. $0.94$).}
\end{table*}

\section{Conclusion}

$F_1$ scores for TF-based unsupervised tokenization—for English and Russian, especially—appear high enough for this technique to inform future experiments in self-reinforcement learning or interpretable unsupervised grammar/language learning.

A new state-of-the art (SOTA) baseline for unsupervised tokenization has been introduced. This baseline may be further reinforced by increasing the complexity and richness of the test corpora.

Optimal thresholds and offshoots of the TF metric vary by language. The process and policy of their discovery and adjustment in an unsupervised manner should be further explored.

Hybridization of TF-based tokenization approaches with lexicon-based ones might be efficient for low-resource and domain-specific languages.

Further unsupervised grammar learning experiments, advancing earlier studies such as those by \citet{5} and \citet{6}, can be run on the basis of our proposed unsupervised tokenization framework. 

Using TF-based segmentation to identify natural boundaries of states and actions for the application of “global feedback” may be explored in the context of reinforcement or experiential learning environments such as in \citeposs{14} work, including ones with delayed/sparse reward.

\section*{Limitations}

The following limitations are known and should be considered when applying the results of this work or relying on them in future studies:

\begin{itemize} 

\item In some cases, tokenization with $N=1$ will not work (e.g., decimal points and dots in web addresses are used as token boundaries). This might be improved with $N>1$, but given the slightly worse performance of such a setup on the explored test set, further studies are needed. Potentially, the notion of “broad tokenization context” (a variant of \citeposs{1} “attention” concept) should be introduced to scale the proposed technology when dealing with richer test corpora.

\item The test corpus of $100$ sentences covers a quite limited subject domain (personal finance), so evaluation on larger and richer corpora is recommended for further studies and applications.

\item While the use of unsupervised parsing based on MI has been found impractical, no full-scale evaluation of this approach has been performed, so no firm claim of its futility can be made; it should be explored and verified further.

\item While the use of TF “peak values” has been explored, no similar “peak values” have been systematically tried for CP \citep{11,12}. Cursory checks rendering low performance of CP offshoots for English and Russian tests indicate that CP “peak values” will not be useful, but a more systematic study is needed for final confirmation.

\item The lack of available memory (32G) made it impossible to explore $N>3$ for the smaller Chinese training corpus and $N>2$ for the larger corpus. While the smaller Chinese corpus has shown $N=2$ providing higher scores compared to $N=1$, and the larger Chinese corpus has shown $N=2$ to be likewise superior, it would be best to confirm this by testing tokenization with $N=3$ on the larger corpus with a $>$32G memory limit.

\item The authors’ lack of Chinese knowledge has prevented reliable interpretations of and judgments regarding the tokenization $F_1$ score, so further exploration involving Chinese tokenization might be required for a more reliable assessment of the presented study’s applicability to the Chinese language.

\end{itemize}

\section*{Ethics Statement}

The presented work appears to have an immediate ethical benefit, due to its contribution to increased inclusiveness in respect to cultures relying on so-called “low-resource” languages and dialects which cannot easily be studied via contemporary linguistic approaches. Presumably, the proposed technology might simplify the study of such languages, providing initial lexicon dictionaries based on raw field data and thereby opening the way for further studies of these languages and their grammars. 

The other long-term positive ethical impact is associated with the “interpretable” nature of this work. Our model contributes to the movement towards open, transparent, and human-friendly linguistic models that can be developed for any human language and delivered to production, thereby precluding “black-box” NLP models from potentially decreasing quality of life.

No negative ethical impacts appear to be connected with this work.   

\section*{Acknowledgments}

We are grateful to Ben Goertzel and Linas Vepstas, who provided us with the initial motivation to work in the Interpretable Natural Language Processing (INLP) domain. We would also like to thank Andres Suarez, who helped us collect some of the training data, and Nikolay Mikhaylovskiy, for engaging in a thoughtful discussion of the results.

\nocite{*}

\bibliography{anthology,custom}

\begin{thebibliography}{20}
\expandafter\ifx\csname natexlab\endcsname\relax\def\natexlab#1{#1}\fi

\bibitem[{Brown et~al.(2020)Brown, Mann, Ryder, Subbiah, Kaplan, Dhariwal,
  Neelakantan, Shyam, Sastry, Askell, Agarwal, Herbert-Voss, Krueger, Henighan,
  Child, Ramesh, Ziegler, Wu, Winter, Hesse, Chen, Sigler, Litwin, Gray, Chess,
  Clark, Berner, McCandlish, Radford, Sutskever, and Amodei}]{2}
Tom Brown, Benjamin Mann, Nick Ryder, Melanie Subbiah, Jared~D Kaplan, Prafulla
  Dhariwal, Arvind Neelakantan, Pranav Shyam, Girish Sastry, Amanda Askell,
  Sandhini Agarwal, Ariel Herbert-Voss, Gretchen Krueger, Tom Henighan, Rewon
  Child, Aditya Ramesh, Daniel Ziegler, Jeffrey Wu, Clemens Winter, Chris
  Hesse, Mark Chen, Eric Sigler, Mateusz Litwin, Scott Gray, Benjamin Chess,
  Jack Clark, Christopher Berner, Sam McCandlish, Alec Radford, Ilya Sutskever,
  and Dario Amodei. 2020.
\newblock \href
  {https://proceedings.neurips.cc/paper/2020/file/1457c0d6bfcb4967418bfb8ac142f64a-Paper.pdf}
  {Language models are few-shot learners}.
\newblock In \emph{Advances in Neural Information Processing Systems},
  volume~33, pages 1877--1901. Curran Associates, Inc.

\bibitem[{Castillo-Domenech and Suarez-Madrigal(2018)}]{20}
Claudia Castillo-Domenech and Andres Suarez-Madrigal. 2018.
\newblock \href {https://hdl.handle.net/20.500.12380/256408} {Statistical
  parsing and unambiguous word representation in {OpenCog}’s unsupervised
  language learning project}.

\bibitem[{Dodge et~al.(2019)Dodge, Gururangan, Card, Schwartz, and Smith}]{19}
Jesse Dodge, Suchin Gururangan, Dallas Card, Roy Schwartz, and Noah~A. Smith.
  2019.
\newblock \href {http://arxiv.org/abs/1909.03004} {Show your work: Improved
  reporting of experimental results}.
\newblock \emph{Computing Research Repository}, arXiv:1909.03004.

\bibitem[{Friston(2010)}]{13}
Karl Friston. 2010.
\newblock \href {https://doi.org/10.1038/nrn2787} {The free-energy principle: a
  unified brain theory?}
\newblock \emph{Nature Reviews Neuroscience}, 11(2):127--138.

\bibitem[{Glushchenko et~al.(2019)Glushchenko, Suarez, Kolonin, Goertzel, and
  Baskov}]{6}
Alex Glushchenko, Andres Suarez, Anton Kolonin, Ben Goertzel, and Oleg Baskov.
  2019.
\newblock Programmatic {Link} {Grammar} induction for unsupervised language
  learning.
\newblock In \emph{Artificial General Intelligence}, pages 111--120, Cham.
  Springer International Publishing.

\bibitem[{Glushchenko et~al.(2018)Glushchenko, Suarez, Kolonin, Goertzel,
  Castillo, Leung, and Baskov}]{5}
Alex Glushchenko, Andres Suarez, Anton Kolonin, Ben Goertzel, Claudia Castillo,
  Man~Hin Leung, and Oleg Baskov. 2018.
\newblock Unsupervised language learning in {OpenCog}.
\newblock In \emph{Artificial General Intelligence}, pages 109--118, Cham.
  Springer International Publishing.

\bibitem[{Gopalakrishnan et~al.(2022)Gopalakrishnan, Irie, Schmidhuber, and van
  Steenkiste}]{15}
Anand Gopalakrishnan, Kazuki Irie, Jürgen Schmidhuber, and Sjoerd van
  Steenkiste. 2022.
\newblock \href {https://doi.org/10.48550/ARXIV.2203.13573} {Unsupervised
  learning of temporal abstractions with slot-based transformers}.
\newblock \emph{Computing Research Repository}, arXiv:2203.13573.

\bibitem[{Jiang and Li(2018)}]{18}
Junfeng Jiang and Jiahao Li. 2018.
\newblock \href {https://doi.org/10.48550/ARXIV.1809.08390} {Constructing
  financial sentimental factors in {Chinese} market using natural language
  processing}.
\newblock \emph{Computing Research Repository}, arXiv:1809.08390.

\bibitem[{Kearsley(2016)}]{11}
Logan Kearsley. 2016.
\newblock A hybrid approach to cross-linguistic tokenization: Morphology with
  statistics.

\bibitem[{Kolonin(2015)}]{4}
Anton Kolonin. 2015.
\newblock \href {https://doi.org/10.1109/SIBIRCON.2015.7361868} {Automatic text
  classification and property extraction applications in medicine}.
\newblock In \emph{2015 International Conference on Biomedical Engineering and
  Computational Technologies (SIBIRCON)}, pages 133--137.

\bibitem[{Kolonin(2022)}]{14}
Anton Kolonin. 2022.
\newblock Neuro-symbolic architecture for experiential learning in discrete and
  functional environments.
\newblock In \emph{Artificial General Intelligence}, pages 106--115, Cham.
  Springer International Publishing.

\bibitem[{Ramesh and Kolonin(2020)}]{8}
Vignav Ramesh and Anton Kolonin. 2020.
\newblock \href {https://doi.org/10.1109/S.A.I.ence50533.2020.9303220}
  {Interpretable natural language segmentation based on {Link} {Grammar}}.
\newblock In \emph{2020 Science and Artificial Intelligence conference
  (S.A.I.ence)}, pages 25--32.

\bibitem[{Ramesh and Kolonin(2021)}]{9}
Vignav Ramesh and Anton Kolonin. 2021.
\newblock \href {http://arxiv.org/abs/2105.00830} {Natural language generation
  using {Link} {Grammar} for general conversational intelligence}.
\newblock \emph{Computing Research Repository}, arXiv:2105.00830.

\bibitem[{Ramesh and Kolonin(2022)}]{10}
Vignav Ramesh and Anton Kolonin. 2022.
\newblock Unsupervised context-driven question answering based on {Link}
  {Grammar}.
\newblock In \emph{Artificial General Intelligence}, pages 210--220, Cham.
  Springer International Publishing.

\bibitem[{Sun et~al.(2018)Sun, Hendrix, Ma, and Baayen}]{16}
Ching~Chu Sun, Peter Hendrix, Jianqiang Ma, and Rolf~Harald Baayen. 2018.
\newblock \href {https://doi.org/10.3758/s13428-018-1038-3} {Chinese {Lexical}
  {Database} ({CLD})}.
\newblock \emph{Behavior Research Methods}, 50(6):2606--2629.

\bibitem[{Vaswani et~al.(2017)Vaswani, Shazeer, Parmar, Uszkoreit, Jones,
  Gomez, Kaiser, and Polosukhin}]{1}
Ashish Vaswani, Noam Shazeer, Niki Parmar, Jakob Uszkoreit, Llion Jones,
  Aidan~N Gomez, \L{ukasz} Kaiser, and Illia Polosukhin. 2017.
\newblock \href
  {https://proceedings.neurips.cc/paper/2017/file/3f5ee243547dee91fbd053c1c4a845aa-Paper.pdf}
  {Attention is all you need}.
\newblock In \emph{Advances in Neural Information Processing Systems},
  volume~30. Curran Associates, Inc.

\bibitem[{Vepstas and Goertzel(2014)}]{3}
Linas Vepstas and Ben Goertzel. 2014.
\newblock \href {http://arxiv.org/abs/1401.3372} {Learning language from a
  large (unannotated) corpus}.
\newblock \emph{Computing Research Repository}, arXiv:1401.3372.

\bibitem[{Vityaev et~al.(2022)Vityaev, Kolonin, Kurpatov, and Molchanov}]{17}
Evgenii Vityaev, Anton Kolonin, Andrey Kurpatov, and Artem Molchanov. 2022.
\newblock \href {https://doi.org/10.48550/ARXIV.2202.12710} {Brain principles
  programming}.
\newblock \emph{Computing Research Repository}, arXiv:2202.12710.

\bibitem[{Wrenn et~al.(2007)Wrenn, Stetson, and Johnson}]{12}
Jesse~O. Wrenn, Peter~D. Stetson, and Stephen~B. Johnson. 2007.
\newblock An unsupervised machine learning approach to segmentation of
  clinician-entered free text.
\newblock In \emph{Proceedings of the AMIA Annual Symposium}, pages 811--5.

\bibitem[{Yuret(1998)}]{7}
Deniz Yuret. 1998.
\newblock \href {http://arxiv.org/abs/cmp-lg/9805009} {Discovery of linguistic
  relations using lexical attraction}.
\newblock \emph{Computing Research Repository}, arXiv:cmp-lg/9805009.

\end{thebibliography}
\bibliographystyle{acl_natbib}

\appendix

\section{Additional Experiments}

\subsection{Symbol Category Clustering}

\subsubsection{Exploration Methodology}

We have tried to explore the extent to which our TF-based model can be used to identify categories of different symbols. For this purpose, we have performed agglomerative clustering of symbols into a dendrogram based on the similarity of symbols (N-grams with $N=1$) stored in the model in the vector space of their adjacent transitions both in forward and backward transitions, based on Cosine and Jaccard similarity measures.

\subsubsection{Experimental Results}

Symbol category clustering experiments have shown a general ability to identify proper groups of English and Russian symbols and letters as well as universal language-agnostic punctuation marks. It is interesting that, even using the Russian corpus, the model was able to properly categorize English letters. It is even more interesting that the symbol category trees for English, when obtained while relying on the Russian corpus, had a cleaner separation of vowels and consonants into individual tree branches, as well as cleaner categorization of punctuation marks (like opening/closing brackets and quotation marks), digits, etc. This can be probably explained by the increased “cleanness” of the English texts embedded in the Russian texts (recall that the best unsupervised tokenization results for English reported above were obtained on the smallest Brown corpus). Both the Cosine and Jaccard similarity measures delivered  similar results, while the categorical trees based on the Jaccard measure appeared more well-balanced and reliable.

\begin{figure}
  \includegraphics[width=0.49\textwidth]{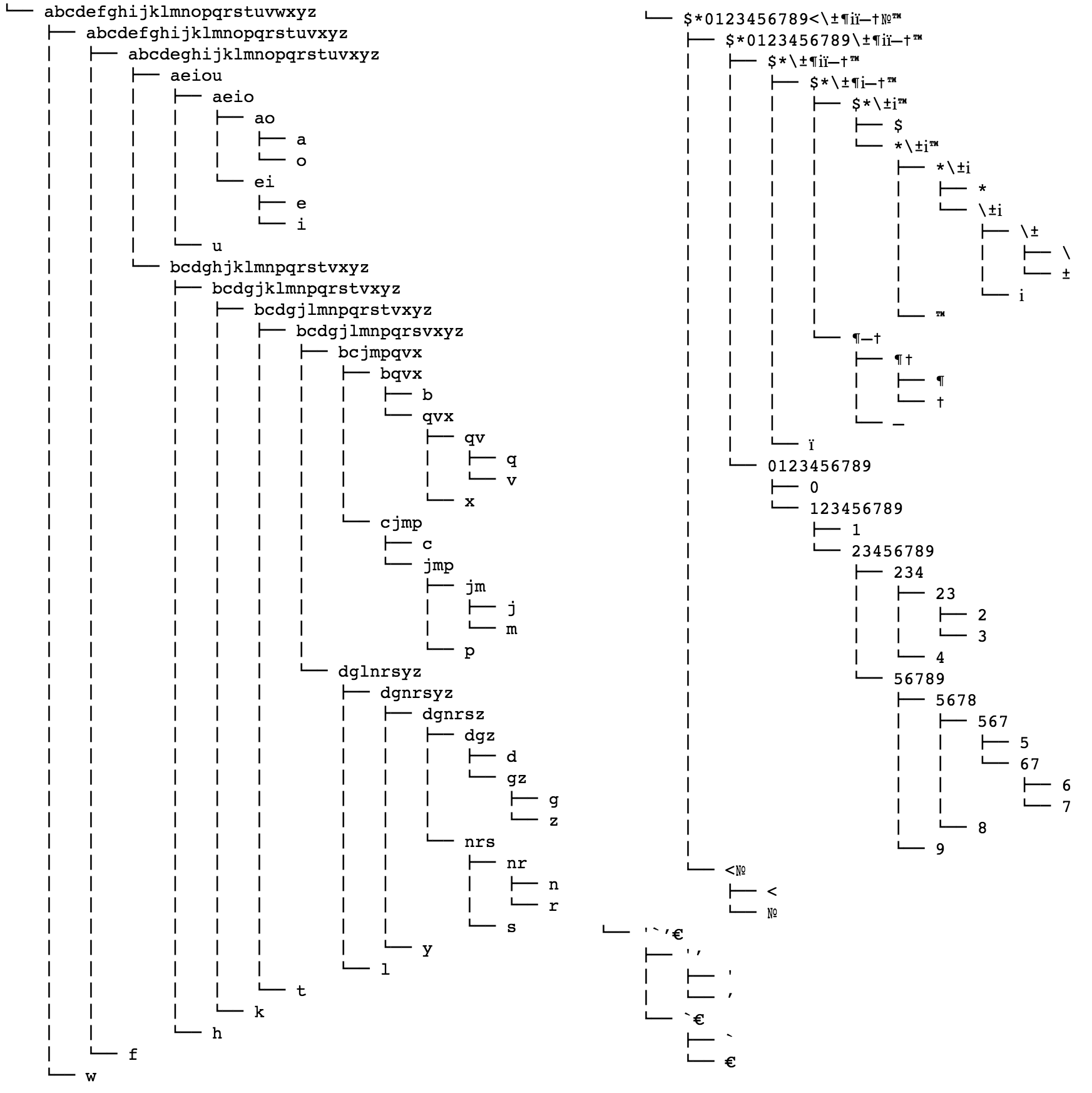}
  \caption{Symbol category agglomerative clustering trees using Jaccard similarity based on the RusAge corpus identifying vowels, consonants, digits, and punctuation mark groups.}
\end{figure}

\subsubsection{Conclusion}

Clustering of parts of speech may provide insights on the morphology and punctuation structure of low-resource and domain-specific languages.

\subsection{Spaceless (“Fluent”) Text Segmentation}

\subsubsection{Exploration Methodology}

We ran tokenization experiments on the input English and Russian texts with white spaces removed to understand the limits of our approach given corpora with continuous (“fluent”) text or speech with no regular and explicit punctuation (as is the case for all Chinese text).

\subsubsection{Experimental Results}

\paragraph{English}

Unsupervised tokenization on fluent text resulted in $F_1=0.42$, while lexicon-based tokenization on the same text yielded $F_1=0.79$ (comparable to the $F_1$ score of $0.82$ on Chinese text) if obtained with search driven by the product $\gamma$ of token length and the logarithm of token frequency. Such results can be explained by the lack of emphasis, articulation, and pauses in spoken communications. In the case of dictionary-based tokenization, as expected, results can be improved by concurrently constructing an alternative tokenization tree that maximizes $\gamma$ across the entire tree, as is being done in Link Grammar and MST Parser \citep[see][]{3, 5, 6} in the case of phrase structure parsing at the sentence level.

\paragraph{Russian}

Unsupervised tokenization on fluent text resulted in $F_1=0.26$, while lexicon-based tokenization on the same text yielded $F_1=0.72$, if obtained with search driven by $\gamma$. The same comments as with spaceless tokenization in English apply here.

\section{Reproducibility}

\subsection{Summary}

The following summary addresses the items from \citet{19} and Joelle Pineau’s reproducibility checklist.

\textbf{A clear description of the mathematical setting, algorithm, and/or model:} the models described by \citet{11} and \citet{12} were extended and used as described in \autoref{sec:s1} of the paper.

\textbf{Source code, with specification of all dependencies, including external libraries:} All source code is contained in the Aigents/Pygents open source project (\url{https://github.com/aigents/pygents}), with usage instructions contained in the following sections of Appendix A. 

\textbf{Description of computing infrastructure used:} MacBook Pro 2018, $2.9$ GHz Intel Core i9 Processor, $32$ GB $2400$ MHz DDR4, Macintosh HD $2$TB.

\textbf{The average runtime for each model or algorithm (e.g., training, inference, etc.), or estimated energy cost:} Model building took between $1$ and $11$ hours per corpus, corresponding to $300$-$3300$ watts of energy consumption. Tokenization for each hyperparameter search trial for a single metric and 3D grid of $3$ parameters was less than $2$ hours, corresponding to $600$ watts.

\textbf{Number of parameters in each model:} Chinese: $143$M and $250$M. English: $12$M and $45$M. Russian: $29$M and $208$M. For each language, the first number corresponds to the smaller corpus, and the second number corresponds to the larger corpus.

\textbf{Corresponding validation performance for each reported test result:} Cross validation was performed in two different yet complementary ways. First, we evaluated different models built upon different independent data sets (smaller and larger corpora) against the same test set independent from the models. In this kind of validation scenario, the best tokenization metrics and N-gram ranks for the best-performing configurations were the same across different corpora and splits for a specific language; model compression thresholds in the range $[0,0.01]$ provided less than $2$\% variance for the best-performing configurations across corpora and splits for a specific language; and the best-performing tokenization threshold for a specific language provided less than $3$\% variance. Second, as mentioned in \autoref{35} of the main paper body, we performed tokenization on the same model with the original test set of $100$ sentences split into two independent subsets of $50$ sentences each; in this kind of validation scenario, the difference in $F_1$ scores for the best-performing configurations of hyperparameters turned out to be less than $1$\% (for English, we had $F_1 = 0.99$ for both test subsets).

\textbf{Explanation of evaluation metrics used, with links to code:} Descriptions of the evaluation metrics can be found in \autoref{s2} of the paper. $F_1$ score assessments were performed using the \texttt{evaluate\_tokenizer\_F1} function (available in \url{https://github.com/aigents/pygents/blob/main/pygents/token.py}) relying on set-based $F_1$ assessment (i.e., calculating the harmonic mean of precision and recall) performed by the \texttt{calc\_F1} function (\url{https://github.com/aigents/pygents/blob/main/pygents/util.py}). 

\textbf{The exact number of training and evaluation runs:} Each training set (smaller corpus, larger corpus, and splits of both) was given a single “clean” run, with a certain number of trial/debugging runs before the final one.  

\textbf{Bounds for each hyperparameter:} $N$: $[1,7]$. Model compression threshold: $[0,0.1]$. Tokenization threshold: $[0,0.9]$.

\textbf{Hyperparameter configurations for the best-performing models:} Reported in subsections \ref{41}, \ref{42}, and \ref{43} in the body of the paper.

\textbf{Number of hyperparameter search trials:} The goal of the presented study was to find the top $F_1$ values possible for completely unsupervised tokenization along with the best-performing hyperparameters, so each unique combination of training corpus, test corpus (or subset of it), and combination of hyperparameters was given exactly one final “clean” run (not counting a certain number of trial/debugging runs before the final one).

\textbf{The method of choosing hyperparameter values (e.g., uniform sampling, manual tuning, etc.) and the criterion used to select among them (e.g., accuracy):} We employed 3D grid search to find the best configurations of hyperparameters, with grid parameters adjusted based on the tokenization $F_1$ scores produced by experimental trial/debugging runs.  

\textbf{Summary statistics of the results (e.g., mean, variance, error bars, etc.):} Top results are shown in \autoref{table}. The variance of $F_1$ scores within intervals of best-performing hyperparameters across corpora was under $3$\%.  

\textbf{For all datasets used, relevant details such as languages, and number of examples:} Reported in \autoref{data} of the paper.

\textbf{Details of train/validation/test splits:} Reported in \autoref{35} of the paper.

\textbf{Explanation of any data that were excluded, and all preprocessing steps:} No data were excluded. Preprocessing steps are briefly covered in \autoref{data} of the paper and explained with more detail in the following section of Appendix A. 

\textbf{Data or link to a downloadable version of the data:} All data and links to data are contained in the accompanying \texttt{data} .zip archive, with usage instructions contained in the following section of Appendix A.

\subsection{Obtaining the Corpora}

\subsubsection{English Training Data}

The Brown training corpus ($6$M size) was downloaded from \url{http://www.sls.hawaii.edu/bley-vroman/brown_nolines.txt} ($6026059$ bytes,  $19810$ lines). For extra validation purposes not presented in the paper, we have used random subsets of $100$ sentences selected from the Brown corpus.

The Gutenberg Children training corpus ($29$M size) was obtained from \url{https://www.gutenberg.org}, based on the books used in the Babi CBT corpus (\url{https://research.fb.com/downloads/babi}). As in \citeposs{20} work, we downloaded the books’ raw text from UTF8 links such as \url{https://www.gutenberg.org/cache/epub/35688/pg35688.txt}, without their original formatting.

The Gutenberg Adult training corpus ($140$M size) was obtained from \url{https://www.gutenberg.org}, based on the selection of $361$ Gutenberg project books with IDs in the range $[53000,53499]$. Once again, raw text was downloaded manually from UTF8 links such as \url{https://www.gutenberg.org/files/53000/53000-0.txt} without formatting.

\subsubsection{Russian Training Data}

Training corpora was downloaded from \url{https://www.kaggle.com/datasets/oldaandozerskaya/fiction-corpus-for-agebased-text-classification}. The two enclosed folders, Test ($141$M size) and Previews ($825$M size), were used independently as alternative training corpora. In further discussion the corpora can be referred to as RusAge Test and RusAge Previews, respectively.

\subsubsection{Chinese Training Data}

The CLUE Benchmark News 2016 dataset was downloaded from \url{https://github.com/brightmart/nlp_chinese_corpus}. When downloaded, the folder \texttt{new2016zh} will have two files, \texttt{news2016zh\_valid.json} ($283711020$ bytes) and \texttt{news2016zh\_train.json} ($8930014780$ bytes), corresponding to smaller and larger training datasets in the scope of our work, respectively. Each of the two files was processed programmatically (parsing JSON; selecting \texttt{title}, \texttt{desc}, and \texttt{content} fields; and saving each of the fields as individual lines), so two plain text files were produced: \texttt{news2016zh\_valid.txt} ($269553996$ bytes, $230391$ lines) and \texttt{news2016zh\_train.txt} ($8481842006$ bytes, $7292256$ lines). In further discussion these corpora can be referred to as “CLUE News 2016 Valid” and “CLUE News 2016 Train,” respectively. (Note that both corpora were used as training datasets, irrespective of their names.)

\subsubsection{Test Data}

The parallel Chinese/English corpus of $100$ multi-sentence statements related to personal finance can be downloaded from Magic Data (\url{https://magichub.com/datasets/chinese-english-parallel-corpus-finance}). It is a tab-delimited text file with individual columns for Chinese and English versions, entitled \texttt{zh} and \texttt{en}, respectively. The Russian extension to it, with only one column entitled \texttt{ru} containing the Russian translations, is contained in the file (\url{https://github.com/aigents/pygents/blob/main/data/corpora/Russian/magicdata/zh_en_ru_100/CORPUS_ZH_EN_RU.txt}) or Aigents/Pygents open source project project.

\subsubsection{Reference Lexicons}

Reference lexicon dictionaries for English and Russian are available as text files from Aigents/Pygents open source project - English: \url{https://raw.githubusercontent.com/aigents/aigents-java/master/lexicon_english.txt}, Russian \url{https://raw.githubusercontent.com/aigents/aigents-java/master/lexicon_russian.txt}.

Reference lexicon dictionaries for Chinese can been downloaded from the following sources: Chinese Lexical Database, or CLD (\url{http://www.chineselexicaldatabase.com/download.php}) \citep[see][]{16};  BLCU Chinese Corpus, or BLC (\url{https://www.plecoforums.com/threads/word-frequency-list-based-on-a-15-billion-character-corpus-bcc-blcu-chinese-corpus.5859}); and SUBTLEX-CH (\url{http://crr.ugent.be/programs-data/subtitle-frequencies/subtlex-ch}). Each of these links contains comma-separated or tab-separated files with lists of words representing Chinese lexicons with different attributions. Individual columns corresponding to words were extracted (along with frequencies of those words, if present), and then a unified Chinese lexicon was created.

\subsection{Experimental Environment}

The \texttt{Python3} code used to run the experiments can be obtained from the Aigents/Pygents open source project (\url{https://github.com/aigents/pygents/}). The external dependencies on Python packages are: \texttt{math}, \texttt{copy}, \texttt{pandas}, \texttt{seaborn}, \texttt{matplotlib}, \texttt{html}, \texttt{urllib}, \texttt{abc}, \texttt{pickle}, \texttt{re}, \texttt{jieba}, and \texttt{numpy}. In order to run the following code, four imports are expected, as follows:

\begin{quote}\begin{small}
\begin{verbatim}
from pygents.token import *
from pygents.text import *
from pygents.util import *
from pygents.plot import *
\end{verbatim}
\end{small}\end{quote}

\subsection{Model Building}

To build the model on the CLUE News 2016 Valid and CLUE News 2016 Train corpora for Chinese, the following code has been used to perform line-by line training on a single file using the \texttt{FreedomTokenizer} class from \url{https://github.com/aigents/pygents/blob/main/pygents/token.py}. The number of parameters, corresponding to number of weights or frequency count for either N-grams or transitions between N-grams, was found to be $143129564$ (corresponding to Valid corpus and $N=\{1,2,3\}$) and $249859247$ (corresponding to Train corpus and $N=\{1,2\}$).

\begin{quote}\begin{small}
\begin{verbatim}
max_n = 3 # 2 for larger Train corpus, 
          # 3 for smaller Valid corpus
zh_chars=FreedomTokenizer(max_n=max_n,
 mode='chars', debug=False)
    with open(join(path, <corpus filename>),
              errors='ignore') as f:
        while True:
            line = f.readline()
            if not line:
                break
            zh_chars.train([line])
zh_chars.store(<model file name>)
print(zh_chars.count_params())
\end{verbatim}
\end{small}\end{quote}

To build the model on the Brown corpus for English, the following code has been used to train the model on the entire corpus at once (number of model parameters $= 52502749$).

\begin{quote}\begin{small}
\begin{verbatim}
brown_text = url_text(“http://www.sls.hawaii
                      .edu/bley-vroman/brow
                      n_nolines.txt”)
brown_chars = 
  FreedomTokenizer(<model file name>,
                  max_n=7,
                  mode='chars',
                  debug=False)
brown_chars.train([brown_text])
brown_chars.store(<model file name>)
print(brown_chars.count_params())
\end{verbatim}
\end{small}\end{quote}

To build the model on the Gutenberg Children and Adult corpora for English, the following code has been used to train the model on the entire corpus at once (number of model parameters $= 12321620$ and $44900866$, respectively). 

\begin{quote}\begin{small}
\begin{verbatim}
def tokenizer_train_folder(t,path):
    onlyfiles = [f for f in listdir(path) 
      if isfile(join(path, f))]
    for file in onlyfiles:
        with open(join(path, file),
          errors='ignore') as f:
            lines = f.readlines()
            t.train(lines)
mode = ‘chars’
child_chars = FreedomTokenizer(max_n=7,
                              mode=mode,
                              debug=False)
tokenizer_train_folder(child_chars,
                      <corpus folder
                        name>)
child_chars.store(<model file name>)
print(child_chars.count_params())
\end{verbatim}
\end{small}\end{quote}

Model building is performed via the \texttt{train} function in the \texttt{FreedomTokenizer} class contained in \url{https://github.com/aigents/pygents/blob/main/pygents/token.py}. The \texttt{train} function calls the \texttt{grams\_count\_with\_gram\_freedoms} function residing in the internal module \url{https://github.com/aigents/pygents/blob/main/pygents/text.py}.

Two different kinds of models could be built based on the \texttt{mode} parameter, which could be either \texttt{chars} or \texttt{grams} (corresponding to either \texttt{N-gram-to-N-char} or \texttt{N-gram-to-N-gram}, respectively). 

The same code as for the Gutenberg Children and Adult corpora was used to build the model on the RusAge Test and RusAge Previews corpora (number of model parameters $ = 28998065$ and $207808799$, respectively).

Each phase of model building took up to $1$ hour for smaller corpora and several hours for larger corpora. While building the CLUE News 2016 Valid model the maximum N-gram rank was $N=3$, and while building the CLUE News 2016 Train model, it was $N=2$, due to the given memory limit. Building the largest model (CLUE News 2016 Train, $N=2$) took $11$ hours, which was the maximum training time across all models and languages.

\subsection{Performing Tokenization}

All tokenization experiments were run via the \texttt{evaluate\_freedom\_tokenizer\_options} function (\url{https://github.com/aigents/pygents/blob/main/pygents/token_plot.py}). The primary argument passed to the function is the tokenizer class (\texttt{FreedomBasedTokenizer}), which is supplied with the metrics used for tokenization in forward and backward directions, a list of different combinations of $N$, and a list of tokenization thresholds, as shown in the following example of English tokenization.

\begin{quote}\begin{small}
\begin{verbatim}
test_df=pd.read_csv(os.path.join(path,
  'CORPUS_ZH_EN_RU.txt'),delimiter='\t')
test_texts = list(test_df['en']) 
                          # or ‘zh’/‘ru’
ref_tokenizer = DelimiterTokenizer()
ngram_params = [[1],[2],[3],[4],[5],[6],
                [7],[1,2],[2,3],[1,2,3],
                [1,2,3,4],[4,5,6,7],
                [1,2,3,4,5],
                [1,2,3,4,5,6,7]]
compression_thresholds = [0,0.0001,0.001,
                          0.01,0.1]
tokenization_thresholds = [0.1,0.2,0.3,
                          0.4,0.5,0.6,
                          0.7,0.8,0.9] 
base=FreedomTokenizer(name=<model file 
                            name>,
                      max_n=7, 
                      mode='chars',
                      debug=False)
title = '$F_1$ - Brown ddf- & ddf+'
for filter_threshold in 
  compression_thresholds:
    if filter_threshold > 0:
        model_compress_with_loss(
          base.model,
          filter_threshold
        )
    parameters = base.count_params()
    title="{} filter={} parameters={}"
      .format(title,
              filter_threshold,
              parameters)
    evaluate_freedom_tokenizer_options(
      test_texts,
      ref_tokenizer,
      FreedomBasedTokenizer(base,'ddf-',
                            'ddf+'),
      ngram_params,
      tokenization_thresholds,
      title=title
    )
\end{verbatim}
\end{small}\end{quote}

Hyper-parameters for metrics passed to the \texttt{FreedomBasedTokenizer} class constructor above could be ‘\texttt{p+}’ or ‘\texttt{p-}’ for conditional probabilities in forward and backward directions, ‘\texttt{dp+}’ or ‘\texttt{dp-}’ for derivatives of CP, ‘\texttt{dvp+}’ or ‘\texttt{dvp-}’ for variances of CP, ‘\texttt{f+}’ or ‘\texttt{f-}’ for TFs in forward and backward directions, ‘\texttt{df+}’ or ‘\texttt{df-}’ for derivatives of TF, ‘\texttt{dvf+}’ or ‘\texttt{dvf}-’ for variances of TF, and ‘\texttt{peak+}’ or ‘\texttt{peak-}’ for “peak values” of TF.

For English and Russian, the reference tokenizer \texttt{DelimiterTokeinzer} (\url{https://github.com/aigents/pygents/blob/main/pygents/token.py}) was used for rule-based tokenization (separating words by spaces and detaching any punctuation marks, counting the latter along with spaces and words as individual tokens):

\begin{quote}\begin{small}
\begin{verbatim}
ref_tokenizer = DelimeterTokenizer()
\end{verbatim}
\end{small}\end{quote}

The following combinations of N-gram ranks, model compression thresholds, and tokenization thresholds were used as hyperparameters for English and Russian.

\begin{quote}\begin{small}
\begin{verbatim}
ngram_params = [[1],[2],[3],[4],[5],[6],[7],
                [1,2],[2,3],[1,2,3],[1,2,3,
                4],[4,5,6,7],[1,2,3,4,5],
                [1,2,3,4,5,6,7]]
compression_thresholds = [0,0.0001,0.001,
                          0.01,0.1]
tokenization_thresholds = [0.1,0.2,0.3,0.4,
                          0.5,0.6,0.7,0.8,
                          0.9] 
\end{verbatim}
\end{small}\end{quote}

For Chinese, \texttt{JiebaTokenizer} (available in \url{https://github.com/aigents/pygents/blob/main/pygents/token.py}) was used as a reference tokenizer.

\begin{quote}\begin{small}
\begin{verbatim}
ref_tokenizer = JiebaTokenizer()
\end{verbatim}
\end{small}\end{quote}

The following combinations of N-gram ranks, model compression thresholds, and tokenization thresholds were used as hyperparameters for Chinese.

\begin{quote}\begin{small}
\begin{verbatim}
ngram_params = [[1],[2],[3],[1,2],[2,3],
                [1,2,3]]
compression_thresholds = [0,0.0001,0.001,
                          0.01,0.1]
tokenization_thresholds = [0.0001,0.0005,
                          0.001,0.005,
                          0.01,0.02,0.05,
                          0.1,0.2,0.4,
                          0.8]
\end{verbatim}
\end{small}\end{quote}

All sets of hyperparameters, including metrics based on CP and TF, different N-gram ranks, model compression thresholds, and tokenization thresholds, were applied for different models across all languages against the same test set. 

For additional validation purposes, in order to confirm the reliability of hyperparameters providing the best $F_1$ scores, the same tokenization experiments were run using different splits of the test set (all $100$ lines, first $50$ lines, last $50$ lines) as well as random sets of $100$ lines selected from the Brown corpus, ensuring that the same hyper-parameters were providing the highest $F_1$ scores with close score values. 

Each tokenization trial for an individual pre-built model given the selected tokenization metrics, involving a $3$-dimensional hyperparameter grid search (\texttt{ngram\_params}, \texttt{compression\_thresholds}, and \texttt{tokenization\_thresholds}), took no more than $2$ hours per trial with $1$ hour as an average.

\subsection{Evaluation}

The \texttt{evaluate\_freedom\_tokenizer\_options} function used to run the experiments discussed previously performed $F_1$ score assessments internally by calling the  \texttt{evaluate\_tokenizer\_F1} function (\url{https://github.com/aigents/pygents/blob/main/pygents/token.py}), which calculated the average $F_1$ score across all input test texts by comparing outputs of the evaluated and reference tokenizers.  

Evaluation of lexicon-based tokenization for reference was done by merging the reference lexicon dictionary with a list of conventional punctuation symbols and using the \texttt{LexiconIndexedTokenizer} class (\url{https://github.com/aigents/pygents/blob/main/pygents/token.py}), as shown in the below code. The \texttt{sortmode} variable denotes whether greedy search is based on token length ($0$), token frequency ($1$), or the product of token length and the logarithm of frequency ($2$).

\begin{quote}\begin{small}
\begin{verbatim}
test_df = pd.read_csv(os.path.join(path,
  'CORPUS_ZH_EN_RU.txt'),delimiter='\t')
test_texts = list(test_df['en']) # or ‘zh’
                                 # or ‘ru’
ref_tokenizer = DelimiterTokenizer() 
  # use JiebaTokenizer() for Chinese

# Get raw lexicon list, use respective 
# source lexicons for English/Russian/
# Chinese
en_lex = list(pd.read_csv("https://raw.\
    githubusercontent.com/aigents/\
    aigents-java/master/lexicon_english\
    .txt",
                          sep='\t',
                          header=None,
                          na_filter=False
                         ).to_records(
                             index=False
                          )
             )

# Add delimiters to the list
delimiters = ' \t\n\r\'`"“”+=-_&/|\*()[]
              <>#^@~,;:.!?'
lex = en_lex + [(i, top_weight) for i 
  in list(delimiters)]
en_lex0_tokenizer = 
  LexiconIndexedTokenizer(
    lexicon=lex, sortmode=0, cased=True
  )
en_lex1_tokenizer = 
  LexiconIndexedTokenizer(
    lexicon=lex, sortmode=1, cased=True
  )
en_lex2_tokenizer = 
  LexiconIndexedTokenizer(
    lexicon=lex, sortmode=2, cased=True
  )
print(t,en_lex0_tokenizer.count_params())

# sort by token length
print(evaluate_tokenizer_F1(test_texts,
  del_tokenizer,en_lex0_tokenizer,
  debug=False))

# sort by frequency
print(evaluate_tokenizer_F1(test_texts,
  del_tokenizer,en_lex1_tokenizer,
  debug=False))

# sort by token length and frequency
print(evaluate_tokenizer_F1(test_texts,
  del_tokenizer,en_lex2_tokenizer,
  debug=False))
\end{verbatim}
\end{small}\end{quote}

Calculation of lexicon discovery precision was achieved by passing extra parameters to the \texttt{evaluate\_tokenizer\_F1} function, so that all tokens identified by the evaluated (freedom-based) and reference (lexicon-based or rule-based) tokenizers could be collected. Upon the collection of the actual (evaluated tokenizer) and expected (reference tokenizer) tokens, the precision values of both the actual and expected counts of tokens were computed (in the below code, “relevant” tokens are those present in the lexicon).   

\begin{quote}\begin{small}
\begin{verbatim}
base = FreedomTokenizer(
  name=<model file name>,
  max_n=7,mode='chars',
  debug=False
)
model_compress_with_loss(base.model,
                         0.0001) 
test_tokenizer = FreedomBasedTokenizer(
  base,'dvf-','dvf+') # for English/Russian
test_tokenizer.set_options(nlist = [1], 
  threshold=0.4)  # for English and Russian

expected = {}
actual = {}
tokenization_F1 = 
  evaluate_tokenizer_F1(
    test_texts,
    del_tokenizer,
    test_tokenizer,
    expected_collector=expected,
    actual_collector=actual
  )

expected_count = sum([expected[key] 
  for key in expected])
relevant_count = sum([expected[key] 
  for key in expected if key.lower() 
  in en_lex_delimited_dict])
irrelevant_count = sum([expected[key] 
  for key in expected if not key.lower() 
  in en_lex_delimited_dict])
print(expected_count,
  relevant_count,
  irrelevant_count,
  relevant_count/expected_count,
  (relevant_count)/expected_count)

actual_count = sum([actual[key] 
  for key in actual])
relevant_count = sum([actual[key] 
  for key in actual if key.lower() 
  in en_lex_delimited_dict])
irrelevant_count = sum([actual[key] 
  for key in actual if not key.lower() 
  in en_lex_delimited_dict])
print(actual_count,
  relevant_count,
  irrelevant_count,
  relevant_count/actual_count,
  (relevant_count)/actual_count)
\end{verbatim}
\end{small}\end{quote}

\end{document}